\pdfoutput=1

\documentclass[11pt]{article}

\usepackage[final]{acl}

\usepackage{times}
\usepackage{latexsym}
\usepackage{subcaption}
\usepackage{booktabs}      %
\usepackage{multirow}      %
\usepackage[table]{xcolor} %
\usepackage{graphicx}      %
\usepackage{colortbl}      %
\usepackage{caption}       %
\usepackage{amsmath}       %
\usepackage{makecell}      %
\usepackage{array}
\usepackage{xcolor}
\usepackage{float}

\usepackage[T1]{fontenc}

\usepackage[utf8]{inputenc}

\usepackage{microtype}

\usepackage{inconsolata}

\usepackage{graphicx}
\usepackage{xspace}
\usepackage{hwemoji}
\usepackage{cleveref}

\newcommand{\ours}{\textsc{LEO-Mini}\xspace}
\newcommand{\tokenselection}{\textsc{CoTR}\xspace}
\newcommand{\CMOE}{\textsc{MMoE}\xspace}

\usepackage{newunicodechar}
\usepackage{graphicx}
\usepackage{float}

\title{\ours: An Efficient Multimodal Large Language Model using Conditional Token Reduction and Mixture of Multi-Modal Experts}

\author{Yimu Wang\thanks{Equal contribution}, Mozhgan Nasr Azadani$^{*}$, Sean Sedwards, Krzysztof Czarnecki\\
    University of Waterloo, Canada\\
  \texttt{\{yimu.wang,mnasraza,sean.sedwards,k2czarne\}@uwaterloo.ca} }

\begin{document}
\maketitle

\begin{abstract}
    Redundancy of visual tokens in multi-modal large language models (MLLMs) significantly reduces their computational efficiency. Recent approaches, such as resamplers and summarizers, have sought to reduce the number of visual tokens, but at the cost of visual reasoning ability. To address this, we propose \ours, a novel MLLM that significantly reduces the number of visual tokens and simultaneously boosts visual reasoning capabilities. 
    For efficiency, \ours incorporates \tokenselection, a novel token reduction module to consolidate a large number of visual tokens into a smaller set of tokens, using the similarity between visual tokens, text tokens, and a compact learnable query. For effectiveness, to scale up the model's ability with minimal computational overhead, \ours employs \CMOE, a novel mixture of multi-modal experts module. \CMOE employs a set of LoRA experts with a novel router to switch between them based on the input text and visual tokens instead of only using the input hidden state.  
    \CMOE also includes a general LoRA expert that is always activated to learn general knowledge for LLM reasoning. 
    For extracting richer visual features, \CMOE employs a set of vision experts trained on diverse domain-specific data. To demonstrate \ours's improved efficiency and performance, we evaluate it against existing efficient MLLMs on various benchmark vision-language tasks.

\end{abstract}

\begin{figure}[h!]
    \centering
    \begin{subfigure}[b]{0.9\linewidth}
        \centering
        \includegraphics[width=\linewidth]{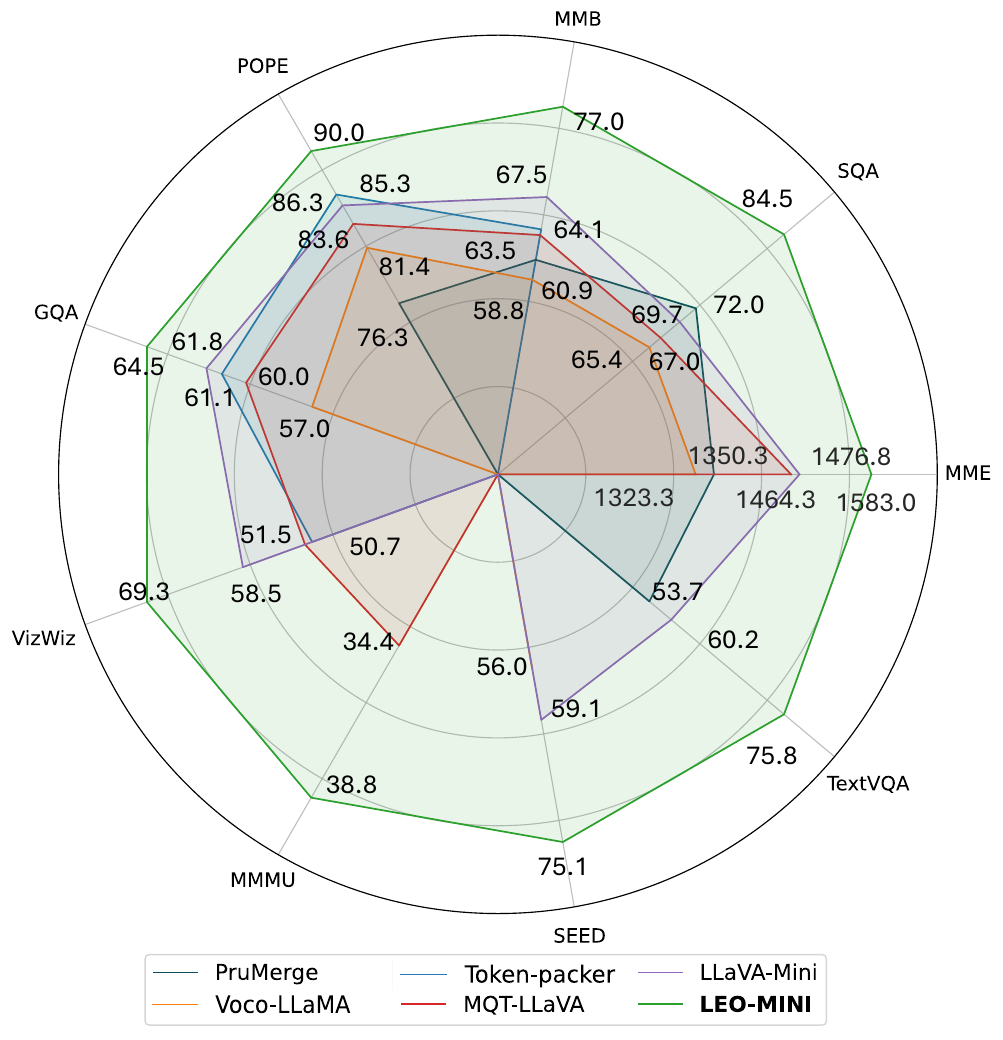}
        \caption{\textbf{Improved effectiveness} compared with token-reduction MLLMs.}
        \label{fig:subfig1}
    \end{subfigure}

    \begin{subfigure}[b]{0.9\linewidth}
        \centering
        \includegraphics[width=\linewidth]{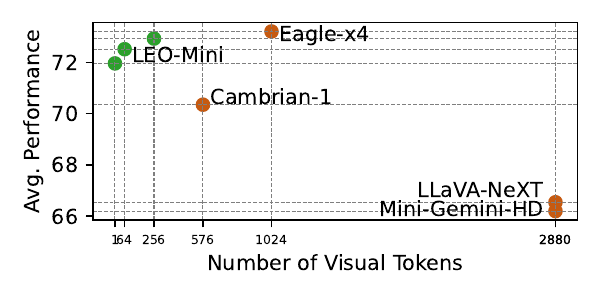}
        \vspace{-2em}
        \caption{\textbf{Improved efficiency} of \ours with general and MoVE-based MLLMs.}
        \label{fig:subfig2}
    \end{subfigure}

    \caption{Performance overview of \ours-Llama3-8B. 
    \subref{fig:subfig1}: Comparison between \ours (64 tokens) and existing token reduction MLLMs %
    , where \ours outperforms all models. \subref{fig:subfig2}: Comparison with other MLLMs
    on 11 vision-language tasks
    using Llama3-8B
    as the LLM. \ours achieves competitive performance while using only 64 visual tokens.}
    \label{fig:teaser}
\end{figure}

\section{Introduction}
\label{sec:intro}

The development of multi-modal large language models (MLLMs)~\cite{azadani2025leo,shi2024eagle,liu2024llavanext,zhang2025llavamini,dai2023instructblip,lu2024deepseek,wang2025hawaiihierarchicalvisualknowledge} has been significantly advanced by aligning vision models~\cite{eva02,Liu_2022_CVPR,pmlr-v202-lee23g} with large-scale pre-trained language models (LLMs)~\cite{grattafiori2024llama3herdmodels,vicuna2023}.
\noindent MLLMs, such as the LLaVA~\cite{liu2024llavanext,liu2024improvedllava1.5}, BLIP~\cite{dai2023instructblip}, and InternVL~\cite{chen2024internvl}, embed image patches into visual tokens through a vision expert~\cite{pmlr-v139-radford21a,Kirillov_2023_ICCV,eva02}.
Then, those visual tokens are input into the LLM for reasoning. 
This has led to strong performance in image and video understanding tasks~\cite{fu_mme_2024,leonardis_mmbench_2025,li_seed-bench_2024,hudson_gqa_2019,lu2022learn,yue_mmmu_2024,li_evaluating_2023,kembhavi_diagram_2016}, bridging the gap between vision and language models.

However, the substantial computational requirement of MLLMs presents a significant challenge to their efficiency. 
In MLLMs, the LLM predominantly drives computational costs, as the vision expert is smaller in comparison. For example, the commonly used vision expert, CLIP-L~\cite{pmlr-v139-radford21a}, has 0.3 billion parameters, whereas LLMs such as LLaMA~\cite{grattafiori2024llama3herdmodels} or Vicuna~\cite{vicuna2023}, have 7--8 billion and 13 billion parameters, respectively. 
While the vision expert is relatively lightweight, its output, i.e., the visual tokens, which is fed into the LLM along with text instruction tokens, significantly increases the computational overhead. 
For instance, CLIP-L~\cite{pmlr-v139-radford21a} encodes a single image into $24 \times 24 = 576$ visual tokens, whereas textual instructions typically consist of fewer than 100 tokens. And this becomes more challenging in high-resolution image understanding~\cite{chen2024internvl,azadani2025leo} or video understanding~\cite{li_videochat_2024,xu_msr-vtt_2016,DBLP:conf/cvpr/HeilbronEGN15}, which requires either more visual tokens per image or processing multiple images.

Reducing the number of visual tokens can thus be an effective strategy for enhancing the efficiency of MLLMs, either through training an efficient compressor~\cite{zhang2025llavamini,li2024tokenpacker} or by using a training-free summarizer~\cite{leonardis_image_2024,zhang_cls_2024,wen_stop_2025}.
In this work, we focus on training-based methods. 
Recent training-based approaches~\cite{zhang2025llavamini,li2024tokenpacker,li2024llama-vid,shang2024llavaprumerge} reduce the number of visual tokens by selecting only the most informative ones~\cite{leonardis_image_2024}, rather than using all visual tokens. Notably, LLaVA-Mini~\cite{zhang2025llavamini} achieves comparable performance to the full LLaVA-1.5~\cite{liu2024improvedllava1.5}, while using only a single visual token. However, aggressively reducing visual tokens may result in the loss of essential visual information, potentially degrading the model's performance.

To extract informative visual features with improved efficiency, in this paper, we propose \ours, a new MLLM that incorporates a novel conditional token reduction module (\tokenselection) for increased efficiency and a novel mixture of multi-modal experts (\CMOE) module for greater effectiveness.

\textbf{Efficiency}.
As the number of visual tokens has become a major bottleneck for MLLM efficiency, \ours introduces \tokenselection to reduce the number of tokens fed into the LLM. 
To focus on the most informative visual tokens based on the input instructions, \tokenselection aggregates visual tokens into a smaller set of consolidated tokens by their similarity to both visual tokens from other vision experts and text tokens. 
Moreover, a learnable query is employed to control the length of consolidated visual tokens, which can be adjusted according to the task or computational requirements. 
This significantly reduces the number of visual tokens, leading to improved training and inference efficiency. 

\textbf{Effectiveness}. 
For a better understanding of visual features and improved reasoning ability, \ours incorporates \CMOE, a novel mixture of multi-modal experts module consisting of \CMOE-LLM and \CMOE-Vision.
Instead of conducting a full finetuning of the entire model after the pretraining, \CMOE-LLM employs a mixture of LoRA experts~\cite{hu2022lora,dou_loramoe_2024,wu_mixture_2023,tian_hydralora_2024} with a novel router and a general expert. 
In contrast to previous work~\cite{hu2022lora,dou_loramoe_2024,wu_mixture_2023,tian_hydralora_2024} whose routers only take the hidden state as input to switch between experts, our router takes the text tokens and the visual tokens as additional input. 
This facilitates more effective switching between different LoRA experts. 
The general expert is continuously activated to learn general knowledge. 
To extract more informative visual features, \CMOE-Vision incorporates multiple vision experts~\cite{pmlr-v139-radford21a,Kirillov_2023_ICCV,eva02}, each trained on data from different domains. 
This boosts the model's ability to understand visual information, leading to improved performance on vision-language tasks while maintaining minimal computational overhead, as both the vision experts and LoRA experts are substantially smaller than the LLM.

Our contributions can be summarized as follows:
\begin{itemize}
    \item
        We propose a new MLLM, \ours, that incorporates a novel token reduction module (\tokenselection) to improve efficiency and a novel mixture of multi-modal experts module (\CMOE) that increases effectiveness. 
    \item
        To the best of our knowledge, \tokenselection is the first to exploit the similarity between visual tokens from multiple vision experts, text tokens, and a small learnable query to focus on the most informative visual tokens.
    \item
        \CMOE-LLM employs a novel router taking visual and text tokens as additional input for better switching between different experts, with a general expert for learning general knowledge. 
        \CMOE-Vision incorporates multiple vision experts for rich visual feature extraction.
    \item 
        We demonstrate the effectiveness and improved efficiency of \ours on various vision-language tasks, as illustrated in \Cref{fig:teaser}. 
\end{itemize}

\section{Related Work}

\textbf{Multi-modal large language models (MLLMs)}. 
As a fundamental problem in multimodal learning~\cite{fang2022multi,fang2023you,fang2024fewer,fang2024not,fang2025turing,wangDeepUnifiedCrossModality2021,wangSearchingPrivatelyImperceptible2020,wang-etal-2023-balance,wangPiecewiseHashingDeep2020,wang-etal-2025-dream,liu-etal-2025-eliot,wang-shi-2023-video,li_evaluating_2023,leonardis_mmbench_2025}, visual understanding has been revolutionized by recent developments in large language models (LLMs). Advancements in LLMs have fueled significant progress in MLLMs, enabling effective cross-modal reasoning through modality fusion and instruction following~\cite{dai2023instructblip,lin2024vila,liu2024improvedllava1.5,li2024monkey,liu2024llavanext}. Early MLLMs struggled with complex visual understanding due to input resolution limits and the inefficiencies of single vision encoders. 
To address this, recent research has enhanced visual experts~\cite{chen2024internvl,zhai2023sigmoid}, incorporated higher-resolution inputs~\cite{li2024mini,luo2024feastllavahr}, and explored mixtures of vision experts~\cite{azadani2025leo,lu2024deepseek,kar2024brave,shi2024eagle,zong_mova_2024,fan2024mousi}. 
Despite the success of these methods, a major challenge remains: the efficiency of MLLMs, as these approaches increase the number of visual tokens, leading to higher computational costs and scalability constraints.

\noindent
\textbf{Compressing visual tokens for MLLMs}. The efficiency of MLLMs is constrained by the LLM backbone’s context length, as high-resolution images generate numerous vision tokens that quickly consume available space and increase computational cost. To address this challenge, recent methods have focused on reducing the number of visual tokens through both training-free~\cite{leonardis_image_2024,huang_dynamic-llava_2024, wen2025stoplookingimportanttokens} and training-based~\cite{zhang2025llavamini,leonardis_image_2024,li2024llama-vid,shang2024llavaprumerge} token reduction strategies. Focusing on training-based approaches, some models aggregate tokens based on visual feature similarities~\cite{shang2024llavaprumerge} or high-low resolution similarities~\cite{li2024tokenpacker}, while others use attention distillation~\cite{ye2024voco-llama}. MQT-LLaVA~\cite{hu2024matryoshka-MQT-llava} employs a query transformer to process a random subset of latent query tokens per step. However, direct compression may lead to information loss. LLaMA-VID~\cite{li2024llama-vid} integrates text tokens as contextual information and applies average pooling for efficient token reduction. 
More recently, LLaVA-Mini~\cite{zhang2025llavamini} mitigates this by combining query-based reduction with prefusion of visual and text tokens.

Existing token reduction approaches~\cite{zhang2025llavamini,li2024llama-vid,hu2024matryoshka-MQT-llava} tend to select tokens by a learnable query or saliency maps. To the best of our knowledge, we propose the first token reduction method that uses text tokens and visual tokens from other visual experts as context to perform attention and reduction over the current vision expert's tokens.

\noindent
\textbf{Mixture of Experts (MoE) in LLMs.} 
MoE is a model design that exploits multiple sparse experts to process different parts of the input space~\cite{jacobs1991adaptive}. Early works~\cite{du2022glam,fedus2022switch} demonstrated that sparse expert activation improves scalability and computational efficiency. Existing MoE-based LLMs~\cite{jiang2024mixtral,dai2024deepseekmoe} typically incorporate MoE by replacing standard feed-forward networks with MoE layers, where each token is routed to a small subset of experts. More recent research~\cite{liu2024pmol,lin2024moe,zadouri2023pushing,wu2024omni} explores integrating MoE with LoRA to further reduce the parameter overhead of traditional MoE models. These methods make use of LoRA’s ability to fine-tune only a small subset of parameters~\cite{wu_mixture_2023,dou_loramoe_2024,chen2024llava}, enabling efficient expert selection and dynamic task adaptation.

In contrast to existing LoRA-based MoEs, which select experts based on the hidden state, our proposed \CMOE-LLM employs a novel routing network that takes visual tokens and textual instructions as additional inputs, enabling expert selection based on multi-modal input. 
Moreover, \CMOE-LLM employs a general expert to learn general knowledge.

\section{Methodology}

In \Cref{sec: model: architecture}, we introduce the overall framework of \ours.
In \Cref{sec: model: token_selection}, we present our proposed token reduction module, \tokenselection, which consolidates a large number of visual tokens into a smaller, more informative set.
Finally, in \Cref{sec: model: mixture_of_experts}, we introduce our mixture of multi-modal experts module, \CMOE, which is designed to enhance the efficiency of fine-tuning MLLMs while preserving their strong performance.

\begin{figure}[t!]
    \centering
    \includegraphics[width=\linewidth]{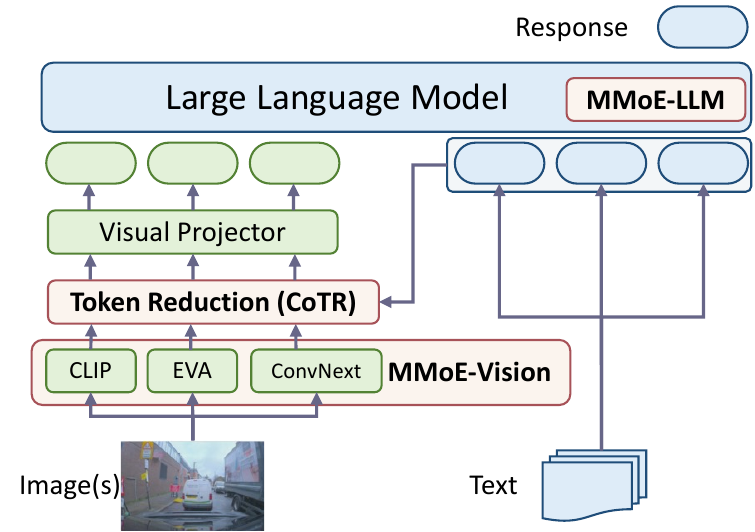}
    \caption{The overview of the proposed \ours. \ours is designed with \CMOE (\Cref{sec: model: mixture_of_experts}) to enhance visual comprehension and \tokenselection (\Cref{sec: model: token_selection}) to reduce the number of visual tokens for efficiency.
    }
    \label{fig: overview}
\end{figure}

\subsection{Architecture}
\label{sec: model: architecture}

The overall architecture of \ours is presented in \Cref{fig: overview}. It follows the general design (vision expert-projector-LLM) of existing MLLMs~\cite{shi2024eagle, azadani2025leo, zhang2025llavamini}, while incorporating a mixture of multi-modal experts and a visual feature compression module.

Specifically, \CMOE introduces \emph{multiple vision experts}, each of which is trained on a domain-specific vision task to extract diverse and informative visual features from the input image. These experts embed the input image into a group of visual tokens $\left\{I_i \in \mathcal{R}^{N_i^V \times d_i^V} \right\}_{i\in[m]}$, where $m$ is the number of vision experts, $N_i^V$ is the number of visual tokens generated by the $i$-th vision expert, and $d_i^V$ is the feature dimension. 

\emph{Visual feature compression} is then applied to the group of visual tokens $\left\{I_i \right\}_{i\in[m]}$. 
First, as the visual tokens generated by different vision experts have different lengths, i.e., $\{N_i^V\}_{i \in [m]}$, the \tokenselection module (\Cref{sec: model: token_selection}) projects them into a group of consolidated visual tokens $\left\{\bar{I}_i \in\mathcal{R}^{N^V\times d^V_i} \right\}_{i\in[m]}$ with the same length of $N^V$. 
$N^V$ is much smaller than $\sum_{i\in[m]}N_i^V$, significantly reducing the number of visual tokens for efficiency. 
Then, the consolidated visual tokens are concatenated channel-wise to form the concatenated visual tokens $\bar{I} \in \mathcal{R}^{N^V \times d^V}$, where $d^V = \sum_{i\in[m]} d_i^V$. 

After that, a \emph{visual projector} is applied to project the concatenated visual token $\bar{I}$ to have the same dimension with the language model input, resulting in $\tilde{I} \in \mathcal{R}^{N^V \times d_{\textsc{LLM}}}$, where $d_{\textsc{LLM}}$ is the feature dimension of the LLM's input.

An LLM $f_{\textsc{LLM}}(\cdot)$ then takes the visual tokens $\tilde{I}$ and the textual instruction tokens $T$ as input to generate the instruction-following response $Y=\{y_i\}_{i\in[L]}$ as,
\begin{equation}
    p(Y|\tilde{I}, T) = \prod_{i=1}^{L} p(y_i|\tilde{I}, T, y_{<i}),
\end{equation}
where $L$ is the length of the response, and $y_{<i}$ is the previous tokens of $y_i$.

\begin{figure}[t!]
    \centering
    \includegraphics[width=\linewidth]{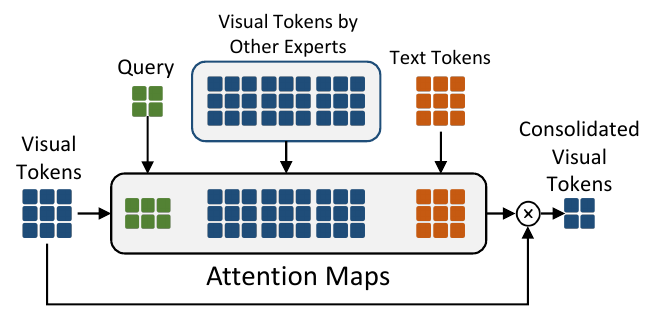}
    \caption{The overview of the proposed \tokenselection. The \tokenselection module takes a group of visual tokens, a learnable query, and text tokens as input, and outputs a consolidated group of visual tokens to reduce the number of visual tokens.}
    \label{fig: token reduction}
\end{figure}

\begin{figure*}[t!]
    \centering
    \includegraphics[width=\linewidth]{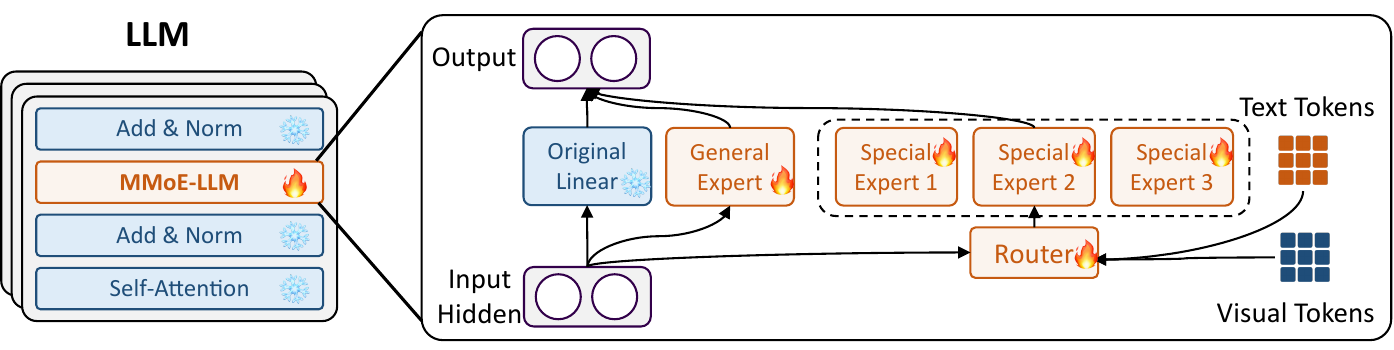}
    \caption{The overview of the proposed \CMOE-LLM for LLM finetuning.
    \CMOE (language) consists of a set of LoRA experts and a routing network that selects the appropriate expert based on the input visual tokens and textual instructions. Moreover, a general expert is employed to learn the general knowledge and improve the robustness of the model.
    }
    \label{fig: cmoe}
\end{figure*}

\subsection{Conditional Token Reduction ({\bf\tokenselection}) }
\label{sec: model: token_selection}
The \tokenselection module, illustrated in~\Cref{fig: token reduction}, is a query-based module that takes a group of visual tokens $\{I_i \}_{i\in[m]}$, text tokens $T$, and a group of query tokens $\{Q_i \in \mathcal{R}^{N^V \times d_i^V}\}_{i\in[m]}$ as input, and outputs a group of consolidated visual tokens $\{\bar{I}_i \}_{i\in[m]}$. 
Specifically, for the visual tokens $I_i$ generated by the $i$-th vision expert, the \tokenselection module computes an attention score $\alpha_i \in \mathcal{R}^{N^V \times N^V_i}$ using the query token $Q_i \in \mathcal{R}^{N^V \times d_i^V}$, text tokens $T$, the visual tokens from other vision experts $\left\{I_j\right\}_{j\in[m]\setminus\{i\}}$, and the visual token $I_i$, as 
\allowdisplaybreaks
\begin{align}
\small
    s_i^{\textsc{query}} =& \hat{Q}_i \hat{I}_i^{\top} \in \mathcal{R}^{N^V \times N^V_i}\,, \\
    s_i^{\textsc{self}} = & \mathbf{1} \hat{I}_i \hat{I}_i^{\top}\in \mathcal{R}^{1 \times N^V_i}\,, \\
    s_i^{\textsc{cross}} = & \sum_{j\in [m]\setminus\{i\}} \hat{I}_j \hat{I}_i^{\top} \in \mathcal{R}^{1 \times N^V_i}\,, \\
    s_i^{\textsc{text}} = & \mathbf{1} \hat{T} \hat{I}_i^{\top} \in \mathcal{R}^{1 \times N^V_i}\,, \\
    \alpha_i = & \frac{\text{softmax}(s_i^{\textsc{query}} + s_i^{\textsc{self}} + s_i^{\textsc{cross}} + s_i^{\textsc{text}})}{\sqrt{d_i^V}}\,,
\end{align}
where $\hat{Q}_i$, $\hat{I}_i, \forall i \in [m]$, and $\hat{T}$ denote the query token, visual tokens, and text tokens projected by learnable linear projections, respectively. 
The term $\mathbf{1}$ is a vector of ones used to compute the self-attention score. 
The attention score $\alpha_i$ is then used to compute the consolidated visual tokens $\bar{I}_i$ as,
\begin{equation}\label{eq: token selection: weighted sum}
    \bar{I}_i = \alpha_i I_i \in \mathcal{R}^{N^V \times d_i^V}\,.
\end{equation}
In this way, redundant visual tokens are aggregated into a compact set of consolidated visual tokens which significantly improves the efficiency. 
The length of the query tokens can be adjusted to control the number of visual tokens according to the specific task requirements. 
Moreover, as we use four different similarities, \tokenselection can capture the complex relationships between multiple sources of features, leading to more informative consolidated visual tokens.

Finally, the consolidated visual tokens $\left\{\bar{I}_i \right\}_{i\in[m]}$ are concatenated channel-wise to form the concatenated visual tokens $\bar{I} \in \mathcal{R}^{N^V \times d^V}$, where $d^V = \sum_{i\in[m]} d_i^V$. 
These concatenated tokens are then projected using a vision projector to have the same dimension as the LLM input size, resulting in $\tilde{I} \in \mathcal{R}^{N^V \times d_{\textsc{LLM}}}$.

\begin{table*}[h!]
  \centering
\resizebox{\textwidth}{!}{%
\begin{tabular}{l|c|ccccc|c|cccc}
\toprule
\multirow{2}{*}{Model} &  \# of&  \multicolumn{5}{c|}{General} & \multicolumn{1}{c|}{OCR} & \multicolumn{3}{c}{Knowledge} \\ 
  &  Visual Tokens & MME$^{P}$ & MMBench & SEED$^{\textit{I}}$ & GQA & VizWiz & TextVQA & SQA & POPE  & MMMU \\
  \midrule
VoCo-LLaMA~\cite{ye2024voco-llama} & 1 & 1323.3 & 58.8 & 53.7 & 57.0  & - & - & 65.4 & 81.4 & - \\
LLaMA-VID~\cite{li2024llama-vid} & 2 & - & - & - & 55.5 & - & 49.0 & 67.7 & 83.1 & - \\
PruMerge~\cite{shang2024llavaprumerge} & 32 & 1350.3 & 60.9 & - & - & - & 56.0 & 72.0 & 76.3 & - \\
MQT-LLaVA~\cite{hu2024matryoshka-MQT-llava} & 64 & 1464.3 & 63.5 & - & 60.0& 51.5 & - & 67.0 & 83.6 & \underline{34.4} \\
Token-Packer~\cite{li2024tokenpacker} & 64 & - & 64.1 & - & 61.1 & 50.7 & - & - & 86.3 & - \\
LLaVA-Mini~\cite{zhang2025llavamini} & 64 & 1476.8 & 67.5 & 60.2 & 61.8 & \underline{58.5} & 59.1 & 69.7 & 85.3 & -\\
\midrule
\rowcolor{green!10} \ours-Vicuna-7B & 64 & \underline{1542.6} & \underline{67.8} & \underline{73.2} & \underline{64.0} & 51.4 & \underline{70.1} & \underline{73.3} & \underline{90.0} & 34.1\\
\rowcolor{green!10} \ours-Llama-8B & 64 & \textbf{1583.0} & \textbf{77.0} & \textbf{75.8} & \textbf{64.5} & \textbf{69.3}  & \textbf{75.1} & \textbf{84.5} & \textbf{90.3} & \textbf{38.8} \\
\bottomrule
\end{tabular}
}
\caption{Comparison to token reduction methods on general, OCR, and knowledge-based tasks. Best in \textbf{Bold}. Second best in \underline{Underline}.}
\label{tab: comparison with token reduction methods}
\end{table*}

\begin{table*}[h!]
\resizebox{\textwidth}{!}{%
\begin{tabular}{l|cccccccccc}
\toprule
Model  & MME$^{P}$ & SEED$^{I}$ & GQA & SQA & MMMU & POPE & AI2D & TextVQA  & ChartQA & OCRBench \\
  \midrule
\rowcolor{green!10} \ours (64 tokens) & {1583} & 75.8 & 64.5 & 84.5 & 38.8 & 90.3 & 75.7 & 75.1 & 80.5 & 62.4 \\
\midrule 
w/ 1 visual token & 1565 & 74.6 & 64.2 & 83.6 & 37.7 & 89.0 & 74.5 & 73.5 & 80.0 & 61.7 \\
w/ 16 visual token & 1577 & 75.4 & 64.4 & 84.3 & 38.2 & 90.1 & 75.4 & 74.2 & 80.2 & \textcolor{teal}{62.6} \\ 
w/ 256 visual token & \textcolor{teal}{1584} & \textcolor{teal}{76.1} & 64.2 & \textcolor{teal}{85.5} & \textcolor{teal}{39.0} & \textcolor{teal}{90.7} & 75.5 & \textcolor{teal}{75.5} & 80.5 & \textcolor{teal}{63.0} \\
\midrule
w/ 1.8m SFT data & 1548 & \textcolor{teal}{76.6} & 64.1 & \textcolor{teal}{85.9} & \textcolor{teal}{39.4} & \textcolor{teal}{90.3} & \textcolor{teal}{77.9} & \textcolor{teal}{75.7} & \textcolor{teal}{80.9} & \textcolor{teal}{63.1} \\
\bottomrule
\end{tabular}
}
\caption{Ablation studies of \ours. We explore the impact of the number of 
visual tokens, the amount of training data, and the \CMOE. Numbers in \textcolor{teal}{green} indicate the performance is improved compared to \ours (64 tokens). }
\label{tab: ablations}
\end{table*}

\subsection{Mixture of Multi-modal Experts ({\bf\CMOE})}
\label{sec: model: mixture_of_experts}
Our mixture of multi-modal experts module, \CMOE, incorporates multiple vision experts to boost visual understanding and multiple LoRA language experts to enhance reasoning. \CMOE comprises \CMOE-Vision and \CMOE-LLM.

\noindent
\textbf{Effective visual comprehension (\CMOE-Vision)}. 
As described in \Cref{sec: model: architecture}, drawing inspiration from previous work~\cite{shi2024eagle,azadani2025leo}, visual tokens are generated by multiple vision experts, each extracting informative features from different perspectives to enrich visual understanding.

\noindent
\textbf{Effective reasoning ability (\CMOE-LLM)}. 
To ensure efficient training with minimal computational overhead, we introduce \CMOE for LLM tuning, following the mixture of LoRA experts~\cite{hu2022lora}, as shown in \Cref{fig: cmoe}. 
The vanilla mixture of LoRA experts consists of a set of experts~\cite{dou_loramoe_2024,tian_hydralora_2024}, \textit{i.e.}, $\{f^E_i(\cdot)\}_{i\in[E]}$, and a routing network $f^{\textsc{routing}}(\cdot)$ that outputs the routing probability $R \in \mathcal{R}^{E}$ taking the hidden state %
as input, where $E$ is the number of experts. 
Then, based on the routing probability, only the top-$1$ expert will be activated.

Different from the vanilla version, the routing network in \CMOE-LLM takes the visual tokens $\tilde{I}$, the textual instruction tokens $T$, and the hidden state $x$ as the input to compute the routing probability $R = \text{softmax}(f^{\textsc{routing}}(\tilde{I}, T, x)) \in \mathcal{R}^{E}$, which facilitates better switching between the experts. 
Moreover, \CMOE-LLM also employs a general expert $f_{\textsc{gen}}^E(\cdot)$ to capture the general knowledge and improve the overall robustness of the model. 

We select $k$ experts with the highest routing probabilities, i.e., $E' = \text{Top}_k(R)$, and compute the output of the \CMOE-LLM as,
\begin{equation}
    \textsc{MMoE-LLM} = f_{\textsc{gen}}^E(x) + \sum_{i\in E'} f^E_i(x) / k\,.
\end{equation}

With the original linear layer $f^{\textsc{ori}}(\cdot)$, the final output is computed as,
\begin{equation}
    \underbrace{f^{\textsc{ori}}(x)}_{\text{Original Linear}} + \underbrace{f_{\textsc{gen}}^E(x)}_{\text{General Expert}} + \underbrace{\sum_{i\in E'} f^E_i(x) / k}_{\text{Selected Experts}}\,.
\end{equation}

\section{Experiments}

\subsection{Implementations and Benchmarks}

\noindent
\textbf{Models}. 
We use Vicuna-v1.5-7B~\cite{vicuna2023} and Llama3-8B~\cite{grattafiori2024llama3herdmodels} as the LLM. 
For vision experts, we follow the general design of EAGLE~\cite{shi2024eagle} and use CLIP~\cite{pmlr-v139-radford21a}, ConvNeXt~\cite{Liu_2022_CVPR}, Pix2Struct~\cite{pmlr-v202-lee23g} and EVA-02~\cite{eva02} for \ours-Llama-8B. 
Similarly, we add another vision expert SAM~\cite{Kirillov_2023_ICCV} for \ours-Vicuna-7B.
The visual projector is a 2-layer MLP with the GELU activation function~\cite{hendrycks2023gaussianerrorlinearunits}. 
\CMOE-LLM is only applied to the MLP in each block of the LLM.
We use 3 experts for \CMOE-LLM with $k=1$ and $1$ general expert being consistently activated. 
Each expert is a LoRA block~\cite{hu2022lora} with rank of $16$. 
The routing network is a 2-layer MLP with the GELU activation function.

More details on training and evaluation are deferred to the Appendix.

\subsection{Main Results}

We compare our \ours with several state-of-the-art token reduction MLLMs (\cref{tab: comparison with token reduction methods}). 
\ours outperforms all the token reduction MLLMs on all tasks. 
Specifically, \ours with Llama3-8B achieves 1583.0 on MME (perception), 77.0 on MMBench, 75.8 on SeedBench, 64.5 on GQA, 69.3 on VizWiz, 75.1 on TextVQA, 84.5 on SQA, 90.3 on POPE, and 38.8 on MMMU, outperforming the best baseline by a large margin. 
This demonstrates the effectiveness of \ours in improving the performance of MLLMs on various tasks. 
Moreover, \ours with Vicuna-7B also outperforms the best baseline on all tasks except for GQA and MMMU, showing the generalization ability of \ours with different LLMs.

\begin{figure*}
    \centering
    \subfloat{\includegraphics[width=0.32\textwidth]{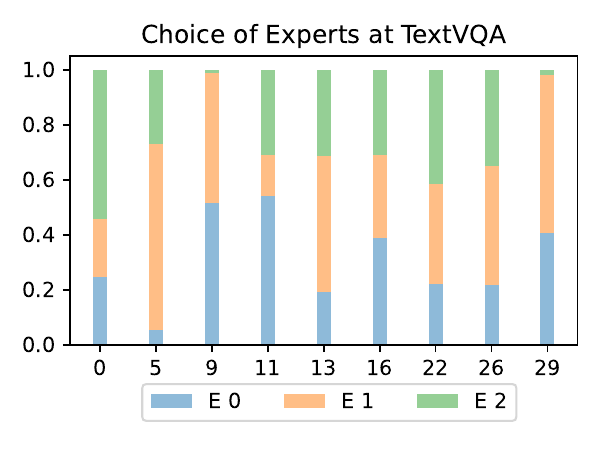}
    }
    \subfloat{\includegraphics[width=0.32\textwidth]{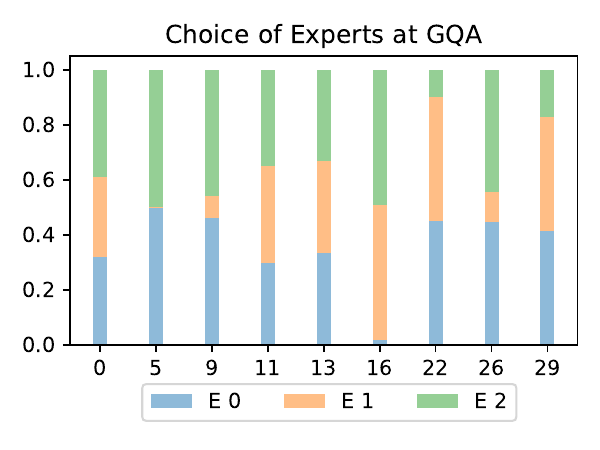}
    }
    \subfloat{\includegraphics[width=0.32\textwidth]{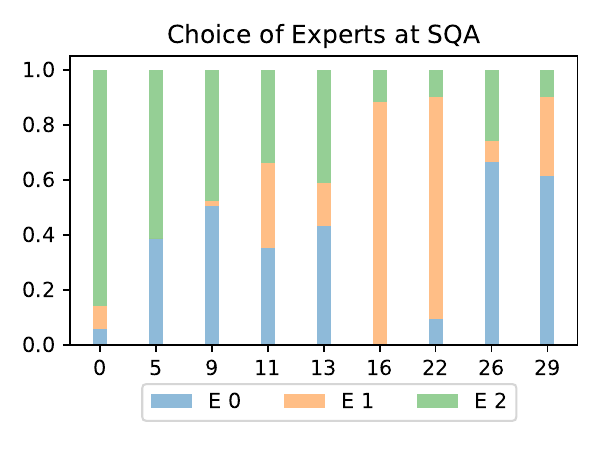}
    }
    \caption{Visualization of the expert choice using \ours-Llama-8B on TextVQA, ScienceQA, and GQA. Best viewed in color. ``E'' refers to Experts.}
    \label{fig: ablation: expert choice}
\end{figure*}

\subsection{Ablation Studies}

In this section, we conduct ablation studies to analyze the effectiveness of different components in \ours, the number of visual tokens, and the amount of SFT data using \ours-Llama-8B. 
Ablations using other token reduction modules and MoE are presented in the Appendix.

\noindent
\textbf{Is 1 visual token enough for MLLMs?}
To understand whether 1 visual token is enough for representing the visual information, we conduct the experiments and present the results in \Cref{tab: ablations}. 
\ours with 1 visual token ($N^V=1$) shows a slight decrease across some metrics compared to \ours with 64 tokens, which is reasonable. 
For example, in the MME$^{P}$, the score dropped from 1583 to 1565, and in SEED$^{I}$, the performance slightly decreased from 75.8 to 75.6. 
Similarly, in the domains of GQA and SQA, the scores decrease from 64.5 to 64.2 and 84.5 to 83.6, respectively. 
This trend continues across other evaluated areas such as MMMU, POPE, AI2D, TextVQA, ChartQA, and OCRBench.
While the use of only 1 visual token consistently leads to lower performance, it greatly improves model's efficiency as shown in \Cref{tab: efficiency}.
To further understand how increasing the number of visual tokens impacts the performance, we conduct experiments with 16 and 256 visual tokens. 
The results show that the performance on most of the benchmarks is improved as the number of visual tokens increases.
This is reasonable as more visual tokens can provide more detailed visual information to the model, which can help the model to better understand the visual information. 
However, we also observe that the performance on some benchmarks is slightly decreased, e.g., AI2D, which might be due to the overfitting issue.

\begin{table*}[h!]
  \centering
  \small
  \resizebox{\textwidth}{!}{%
\begin{tabular}{l|l|c|ccccccccccccc}
\toprule
    {LLM} &Model &  {\# V tokens} & {MME$^{P}$} &   {MMBench} & {SEED$^{I}$} & {GQA} & {SQA} & {MMMU}&{POPE} & {AI2D} & {VizWiz} & { TextVQA }  & { DocVQA } & {ChartQA}  & {OCRBench} \\
        \midrule
\multirow{14}{*}{\rotatebox{90}{Vicuna-7B\& Qwen-7B}} & 
    InstructBLIP~\cite{dai2023instructblip} & 32 & - & - & - & 49.2 & 60.5 & - & - & - & 34.5 & 50.1 & - & - & - \\
    & LLaVA-1.5~\cite{liu2024improvedllava1.5} & 576 & 1511 & 64.3 & 66.1 & 62.0 & 66.8 & - & 85.9 & - & 50.0 &  58.2 & - & - \\
    & LLaVA-NeXt~\cite{liu2024llavanext} & 2880 & 1519 & - & - & 64.2 & 70.1 & 35.1 & - & 66.6 & 57.6 & 64.9 & 74.4 & 54.8 & - \\
    &InternVL~\cite{chen2024internvl} & 1792 & 1525 & 64.3 & - & 62.9 & - & - & 86.4 & - & 52.5 & 57.0 & - & - & - \\
    & VILA~\cite{lin2024vila} & 576 & 1533 & 68.9 & 61.1 & 62.3 & 68.2  & - & 85.5 & - & 57.8 & 64.4 & - & - & - \\
    & Monkey~\cite{li2024monkey} & 256 & - & - & - &  60.7 & 69.4 & - & - & 62.6 & \textbf{61.2} & 67.6 & 66.5 & 65.1 & - \\
    & \multicolumn{15}{c}{\cellcolor{gray!20} MoVE-based MLLMs} \\
    & LLaVA-HR~\cite{luo2024feastllavahr} & 1024 & \textbf{1554} & - & 64.2 & 64.2 & 65.1 & - & 87.6 & - & 48.7 & 67.1 & - & - \\
    & Brave-X5~\cite{kar2024brave} & 160 & - & - & - & 52.7 & & - & 87.6 & - & 54.2 & -  & - & - & - \\     
    & Mini-Gemini~\cite{li2024mini} & 576 & 1523 & 65.8 & - & 64.5 & {71.1} & 36.1 & - & - & - & 65.2 & - & - & - \\
    & Mousi-X3~\cite{fan2024mousi} & 576 & - & 66.8 & 66.0 & 63.3 & 70.2 & - & 87.3 & - & - &  58.0 & - & - & - \\
    & LEO~\cite{azadani2025leo} & 512  & - & \underline{72.9} & 72.2 & \underline{64.8}  & \textbf{78.5} & {36.4} & 88.0 & 69.6 & \underline{57.9} & 68.8 & \textbf{80.1} & \textbf{71.0} & - \\
    & DeepSeek-VL~\cite{lu2024deepseek} & 576 & - & \textbf{73.2} & 70.4 & - & - & \textbf{36.6} & 88.1 & - & - & - & - & - & 45.6 \\
    & Eagle-X5~\cite{shi2024eagle} & 1024 & 1528 & 68.4 & \textbf{73.9} & \textbf{64.9} & 69.8 & 36.3 & \underline{88.8} & - & 54.4 & \textbf{71.2} & \underline{78.6} & \underline{67.8} & \underline{52.9} \\
    \cmidrule(lr){2-16}
     & \cellcolor{green!10} \ours-Vicuna-7B & \cellcolor{green!10} \textbf{64} (\textcolor{teal}{ $\downarrow$ 97.77\%}) & \cellcolor{green!10} \textbf{1543} & \cellcolor{green!10}  {67.8} &\cellcolor{green!10} \underline{73.2} &\cellcolor{green!10} 64.0 &\cellcolor{green!10} \underline{73.3} &\cellcolor{green!10} 34.1 &\cellcolor{green!10} \textbf{90.0} &\cellcolor{green!10} \textbf{72.2} &\cellcolor{green!10} 
     {51.4} &\cellcolor{green!10} \underline{70.1} &\cellcolor{green!10} {75.3} &\cellcolor{green!10} 66.8 & \cellcolor{green!10}\textbf{55.6} \\
    \midrule
    \multirow{6}{*}{\rotatebox{90}{Llama-8B}}
    & Cambrian-1~\cite{tong2024cambrian} & 576 & 1547 & \underline{75.9} & 74.7 & {64.6} & 80.4 & \underline{42.7} & - & 73.0  & - &  71.7 & 77.8 &  73.3 & 62.4  \\
    & LLaVA-NeXT~\cite{liu2024llavanext} & 2880 & \underline{1604} & 72.5 & 72.7 & \underline{65.2} & 72.8 & 41.7 & - & 71.6  & -  & 64.6 & 72.8  & 69.5 & 49.0 \\
    & \multicolumn{15}{c}{\cellcolor{gray!20} MoVE-based MLLMs} \\
    & Mini-Gemini-HD~\cite{li2024mini}  & 2880 & \textbf{1606} & 72.7 & 73.2 & 64.5 & 75.1 & 37.3 & - & 73.5  & - & 70.2 & 74.6 & 59.1 & 47.7 \\
    & Eagle-X4-Plus~\cite{shi2024eagle} & 1024 & 1559 & \underline{75.9} & \textbf{76.3} & \textbf{64.9} & \underline{84.3} & \textbf{43.4} & - & \textbf{76.1}  & - & \textbf{77.1} & \textbf{86.6} & \underline{80.1} & \textbf{62.6}\\
    \cmidrule(lr){2-16}
     & \cellcolor{green!10} \ours-Llama-8B & \cellcolor{green!10} \textbf{64} (\textcolor{teal}{ $\downarrow$ 97.77\%}) & \cellcolor{green!10} 1583 & \cellcolor{green!10}  \textbf{77.0} &\cellcolor{green!10} \underline{75.8} &\cellcolor{green!10}  64.5 &\cellcolor{green!10}  \textbf{84.5} &\cellcolor{green!10}  38.8 &\cellcolor{green!10} \textbf{90.3}  &\cellcolor{green!10} \underline{75.7} &\cellcolor{green!10} \textbf{69.3} &\cellcolor{green!10} \underline{75.1}  &\cellcolor{green!10}   \underline{86.3} &\cellcolor{green!10}\textbf{80.5} & \cellcolor{green!10}\underline{62.4} \\
\bottomrule
\end{tabular}
}
\caption{Comparison to general and MoVE (mixture of vision experts)-based MLLMs. Best in \textbf{Bold}. Second best in \underline{Underline}.}
\label{tab: comparison with general MLLMs}
\end{table*}

\noindent
\textbf{Will more training data help on summarizing the visual information?}
To understand the impact of the amount of training data on the performance, we conduct the experiment with EAGLE-1.8M SFT data~\cite{shi2024eagle} in \Cref{tab: ablations} with 64 tokens.
The results show that the performance is improved on most of the benchmarks, i.e., SeedBench$^{\textit{I}}$, SQA, MMU, POPE, AI2D, TextVQA, ChartQA, and OCRBench. 
For example, the performance on SeedBench$^{\textit{Image}}$ is improved from 75.8 to 76.6, and the performance on SQA is improved from 84.5 to 85.9. 
This indicates that more training data leads to better performance.

\begin{figure*}[t!]
    \centering
    \includegraphics[width=\linewidth]{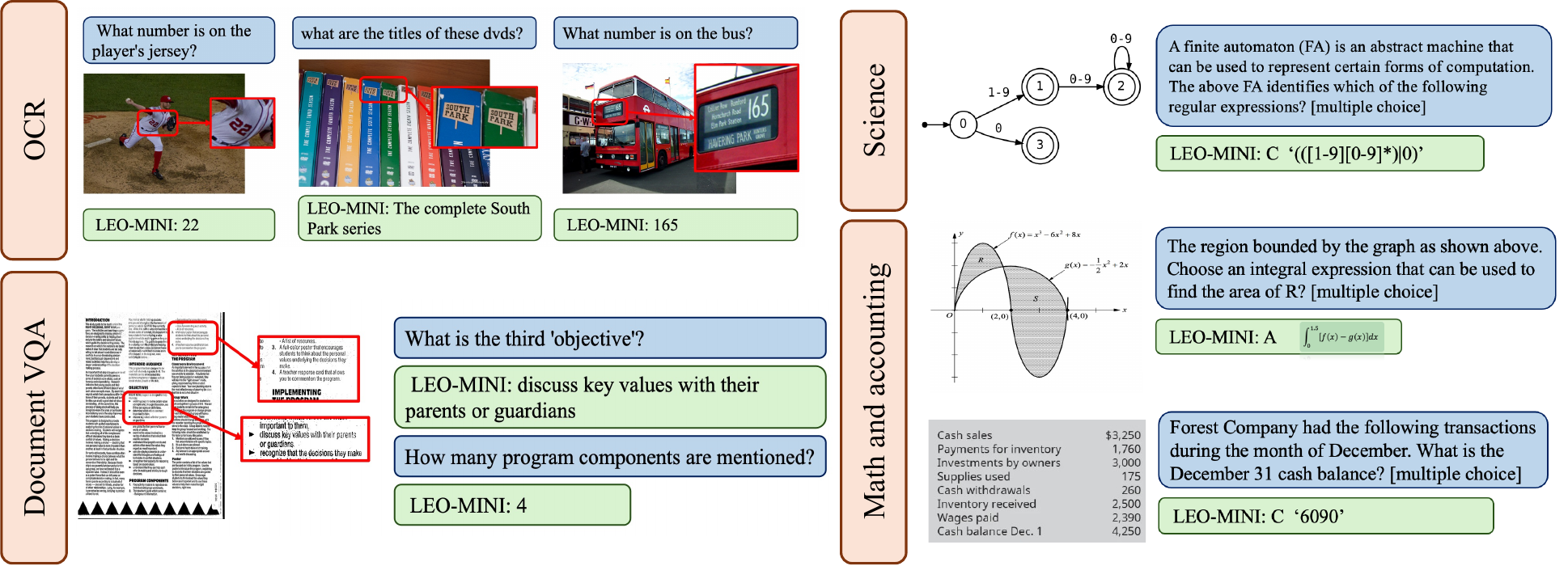}
    \caption{Qualitative results across four vision-language tasks demonstrating LEO-MINI's detailed visual understanding. The images are taken from TextVQA~\cite{singh_towards_2019}, DocVQA~\cite{mathew_docvqa_2021}, and MMMU~\cite{yue_mmmu_2024}.    
    }
    \label{fig: qualitative}
\end{figure*}

\noindent
\textbf{How does \CMOE-LLM switch between different LORA experts?}
We visualize the expert choice in \cref{fig: ablation: expert choice}
with GQA (General), TextVQA (OCR), and SQA (Knowledge). 
We observe that the model effectively switches between different LoRA experts based on the input data. 
For example, in the SQA task, the model mainly activates expert 2 at layer 0, while for TextVQA and 
 GQA, the model evenly activates three experts. 
When it comes to the final layer, such as layer 29, the model evenly activates experts 0 and 1 for TextVQA and GQA, while for SQA, the model mainly activates expert 1, respectively.
This indicates that \CMOE-LLM effectively utilizes the multi-modal input instructions to switch between different experts.
We also compare the expert choice between \textbf{\ours and the vanilla MoE}. 
Results are presented in \Cref{sec: ablation: appendix}.

\subsection{How Does \ours Compare to General MLLMs?}
\noindent
We also compare \ours to general and MoVE-based MLLMs (\Cref{tab: comparison with general MLLMs}).
With Vicuna-7B, \ours outperforms all the general and MoVE-based MLLMs on MME, POPE, AI2D, and OCRBench by 15, 1.2, 2.6, and 2.7 points, respectively, with only 64 visual tokens. 
\ours with Vicuna-7B also achieves the second best performance on SeedBench$^{I}$, SQA, and TextVQA.
On the other side, with Llama3-8B, \ours achieves stronger performance as it outperforms all the general and MoVE-based MLLMs on MMBench, SQA, and CharQA by 1.1, 0.2, and 0.4 points, respectively, while reducing the number of visual tokens by 97.77\% compared to LLaVA-NeXt and Mini-Gemeni-HD.
Moreover, compared to the general MLLMs, \ours with Llama3-8B achieves comparable performance on other benchmark datasets as the discrepancy is marginal. 

\subsection{Qualitative Analysis}

To understand the detailed visual understanding of \ours, we conduct a case study for qualitative analysis on four vision-language tasks~\cite{singh_towards_2019,mathew_docvqa_2021,yue_mmmu_2024} as shown in \Cref{fig: qualitative}. 
We use \ours-Llama3-8B with 64 tokens. 

Though the model only takes 64 visual tokens, \ours performs effectively in capturing visual details, such as accurately identifying the numbers and small book title in OCR. 
Moreover, \ours is able to understand the order and count numbers, as for document VQA, \ours precisely finds the correct results. 
For science and math questions, \ours also shows incredible reasoning ability. 
For science, \ours can effectively translate the computation diagram into mathematical equations. 
For math and accounting, \ours successfully finds the true expression and does calculations correctly.

\subsection{Efficiency Analysis}

To understand how efficient \ours is, we compare the number of visual tokens, FLOPs, and CUDA processing time of \ours with other 
We also compare \ours to MLLMs~\cite{liu2024llavanext,shi2024eagle,tong2024cambrian} using Llama3-8B. 
Specifically, we use \href{https://pytorch.org/docs/stable/profiler.html}{Pytorch Profiler} to measure the FLOPs, CUDA time, and GPU memory usage using \href{https://github.com/haotian-liu/LLaVA/blob/1a91fc274d7c35a9b50b3cb29c4247ae5837ce39/images/llava_v1_5_radar.jpg?raw=true}{the figure} and ``What is shown in this image?'' as input. 
We use one A6000 GPU with 48 GB memory to inference the models.

The comparison is shown in \Cref{tab: efficiency}. 
\ours with Llama3-8B is 97.77\% more efficient in terms of visual tokens, 63.83\% in terms of FLOPs, and 90.69\% in terms of CUDA time compared to LLaVA-Next. 
Moreover, as shown in \Cref{fig:teaser}, \ours outperforms LLaVA-Next by 6\% on average performance with only 64 visual tokens.
Even compared to the most powerful MLLM with similar performance, \textit{i.e.}, Eagle-X4-Plus, \ours is 93.75\% more efficient in terms of visual tokens, 44.10\% for FLOPs, and 68.40\% for CUDA time.
More excitingly and importantly, \ours with only 1 token achieves comparable performance to the state-of-the-art MLLMs with thousands of visual tokens with faster inference time and lower computational cost in terms of FLOPs.
\begin{table}[t!]
\centering
\resizebox{\columnwidth}{!}{%
  \begin{tabular}{l|cccc}
    \toprule
    Models & \# of VT & FLOPs (T) & CUDA Time (s) & GPU Memory (GB) \\
    \midrule
    LLaVA-Next~\cite{liu2024llavanext} & 2880 & 45.5 & 7.037 & 14.74 \\
    Eagle-X4-Plus~\cite{shi2024eagle} & 1024 & 29.4 & 2.073 & 20.10 \\
    \midrule
    \rowcolor{green!10} \ours-Llama-8B & 1 & 13.7 & 0.593 & 19.28 \\
    \rowcolor{green!10} \ours-Llama-8B & 64 & 16.4 & 0.655 & 19.34 \\
    $\Delta$ & \textcolor{teal}{ $\downarrow$ 97.77\%} & \textcolor{teal}{ $\downarrow$ 63.83\%}& \textcolor{teal}{ $\downarrow$ 90.69\%} \\
    \bottomrule
  \end{tabular}%
}
\caption{Efficiency analysis of \ours and other MLLMs using Llama3-8B. ``VT'' represents visual tokens. $\Delta$ indicates the difference between \ours and LLaVA-Next.}
\label{tab: efficiency}
\end{table}

\section{Conclusion}

In this paper, to address the redundancy of visual tokens in MLLMs, we propose a novel MLLM, \ours, that significantly reduces the number of visual tokens while boosting visual reasoning capabilities.
\ours incorporates a novel token reduction module, \tokenselection, to consolidate a large number of visual tokens into a smaller set of tokens, using the similarity between visual tokens, text tokens, and a compact learnable query. 
However, simply reducing the number of visual tokens leads to an information loss. 
To avoid the loss and boost the visual comprehension ability with minimal computational overhead, \ours employs a novel mixture of multi-modal experts module, \CMOE, that includes a set of language (LoRA) experts and a set of vision experts trained on diverse domain-specific data.
For better switching between different LoRA experts, \CMOE employs a new router that takes the text and visual tokens as additional inputs. 
\CMOE also includes a general LoRA expert that is always activated to learn general knowledge. 
We evaluate \ours on various vision-and-language tasks, showcasing its potential for practical applications with improved efficiency and performance compared.

\section{Limitations}
First, the proposed \tokenselection needs training. 
Introducing a training-free token reduction module might be a promising direction.
Second, the proposed \CMOE is designed to be efficient and scalable, but it may not be optimal for all tasks.
For vision experts, due to the computational limitations, we use a fixed set of experts. 
It would be interesting to explore more vision experts from a wide range of domains and introduce more advanced expert selection mechanisms, such as dynamic vision expert routing. 
Last, due to the computational limitations, we did not test the efficiency of \ours using bigger LLMs with 13B or 67B parameters. 

\section{Potential Risks}

\ours is an MLLM that can be used for various tasks, including visual question answering, image captioning, and document understanding.
However, as an inherent risk of (multi-modal) LLMs, \ours may generate biased or harmful content, especially when the query data contains sensitive information. 
We recommend that users carefully review the generated content and apply appropriate filters to mitigate potential risks. 

\section*{Acknowledgement}

This work was supported by the Natural Sciences and Engineering Research Council of Canada (NSERC)-CSE Research Community project entitled ``An End-to-End Approach to Safe and Secure AI Systems'' and NSERC's Postdoctoral Fellowship. 
Researchers funded through the NSERC-CSE Research Communities Grants do not represent the Communications Security Establishment Canada or the Government of Canada. 
Any research, opinions, or positions they produce as part of this initiative do not represent the official views of the Government of Canada.

\bibliography{custom}

\appendix
\label{sec:appendix}

\clearpage
\setcounter{page}{1}
\begin{table*}[h!]
    \centering
    \resizebox{\textwidth}{!}{%
    \begin{tabular}{cc|cccccccccc}
    \toprule
    Token Reduction & MoE & MME$^{P}$ & SEED$^{I}$ & GQA & SQA & MMMU & POPE & AI2D & TextVQA  & ChartQA & OCRBench \\
    \midrule
    \rowcolor{green!10} \tokenselection & \CMOE-Vision + \CMOE-LLM & \textbf{1583} & \textbf{75.8} & \textbf{64.5} & \textbf{84.5} & \textbf{38.8} & 90.3 & \textbf{75.7} & \textbf{75.1} & \textbf{80.5}& \textbf{62.4} \\
    \midrule 
    MQT & \CMOE-Vision + \CMOE-LLM & 1537 & 75.3 & 64.4 & 84.2 & 37.0 & \textbf{90.4} & 75.3 & 74.2 & 80.4 & 61.6 \\
    MQT & - & 1435 & - & 61.6 & 67.6 & 34.8 & 84.4 & - & - & - & - \\
    \midrule
    \tokenselection & \CMOE-Vision + LoRA-MoE & 1521 &75.3 & \textbf{64.5} & 83.4 & 38.2 & 90.3 &75.4& 74.0 & 80.0 & 61.7 \\
    \bottomrule
    \end{tabular}
    }
    \caption{Ablation study on the effectiveness of \tokenselection and \CMOE-LLM. We use Llama3-8B as the backbone.}
    \label{tab: ablations : MQT}
\end{table*}

\begin{figure*}
    \centering
    \subfloat{\includegraphics[width=0.32\textwidth]{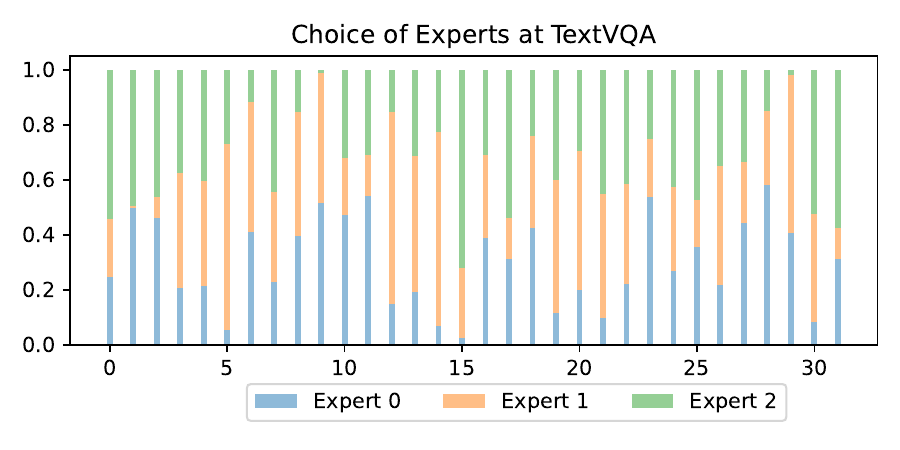}
    }
    \subfloat{\includegraphics[width=0.32\textwidth]{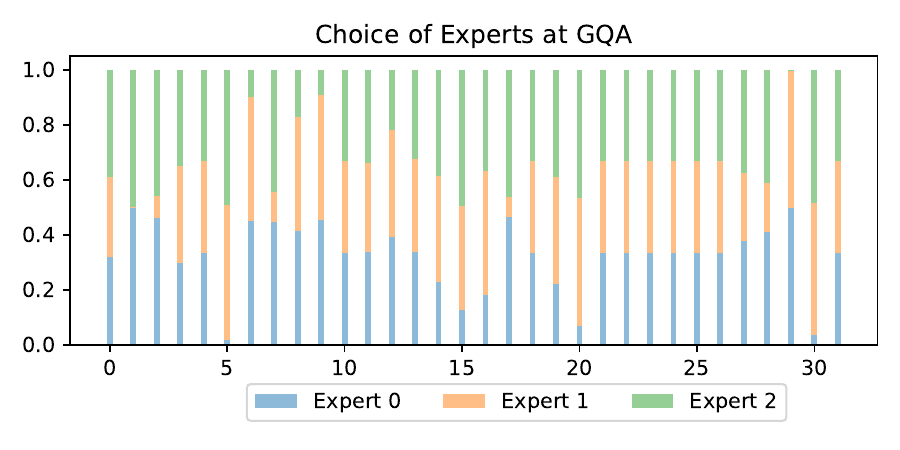}
    }
    \subfloat{\includegraphics[width=0.32\textwidth]{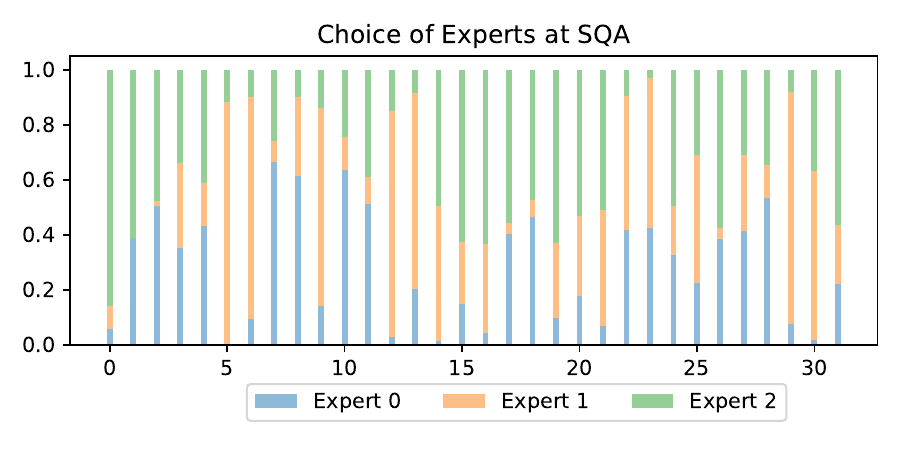}
    }
    \caption{Visualization of the expert choice using \ours-Llama-8B on TextVQA, ScienceQA, and GQA. Best viewed in color. }
    \label{fig: ablation: expert choice full}
\end{figure*}

\begin{figure*}
    \centering
    \subfloat{\includegraphics[width=0.32\textwidth]{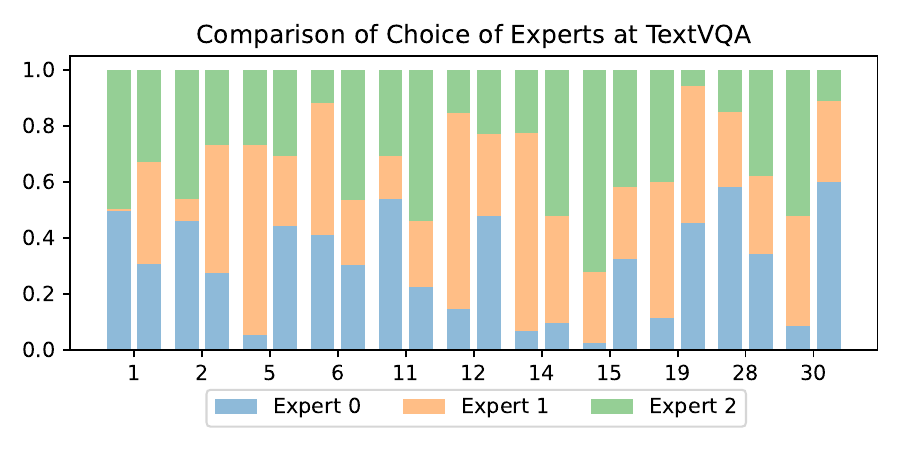}
    }
    \subfloat{\includegraphics[width=0.32\textwidth]{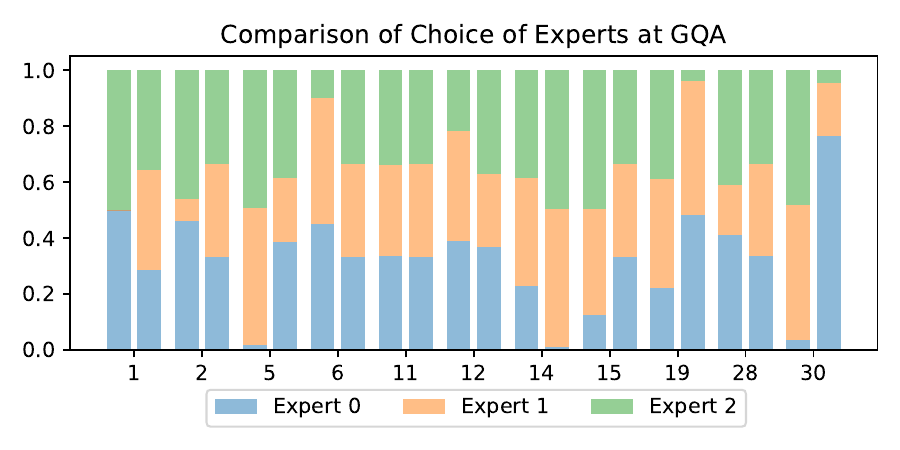}
    }
    \subfloat{\includegraphics[width=0.32\textwidth]{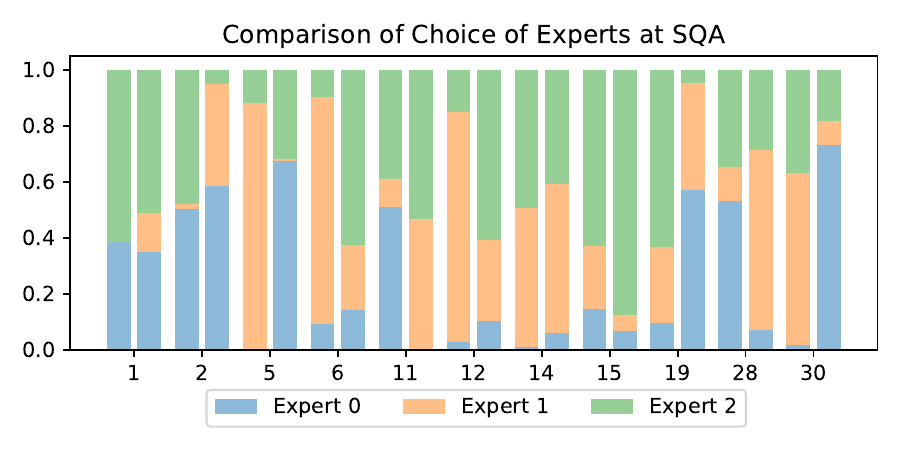}
    }
    \caption{Visualization of the expert choice using \ours-Llama-8B on TextVQA, ScienceQA, and GQA compared to the vanilla MoE. Best viewed in color.}
    \label{fig: ablation: expert choice comparison}
\end{figure*}

\begin{figure*}
    \centering
    \subfloat{\includegraphics[width=0.32\textwidth]{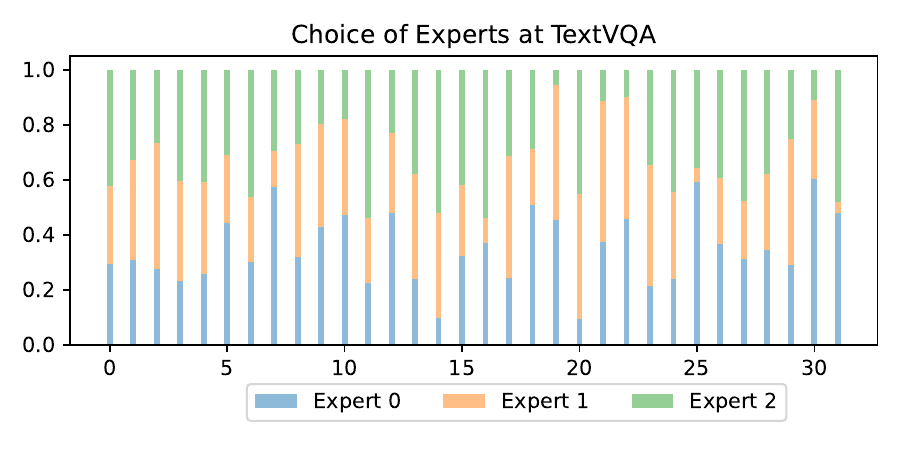}
    }
    \subfloat{\includegraphics[width=0.32\textwidth]{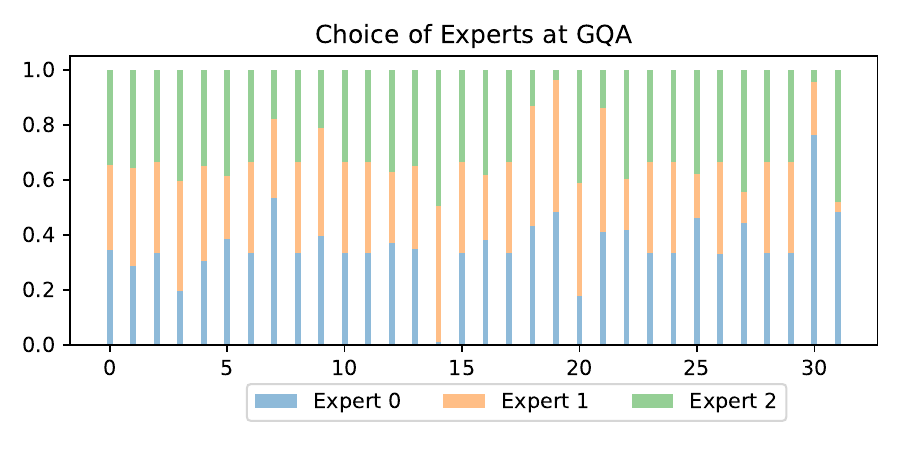}
    }
    \subfloat{\includegraphics[width=0.32\textwidth]{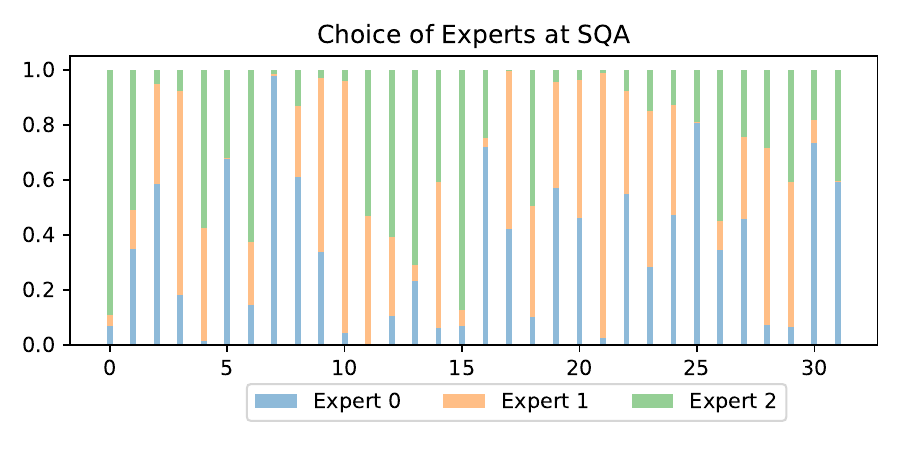}
    }
    \caption{Visualization of the expert choice using the vanilla MoE on TextVQA, ScienceQA, and GQA. Best viewed in color.}
    \label{fig: ablation: expert choice vanilla MoE}
\end{figure*}

\section{Experiments}

\subsection{Benchmark Datasets}

We evaluate our method on the following benchmark datasets: MME~\cite{fu_mme_2024}, MMBench~\cite{leonardis_mmbench_2025}, Seed-Bench~\cite{li_seed-bench_2024}, GQA~\cite{hudson_gqa_2019}, SQA~\cite{lu2022learn}, MMMU~\cite{yue_mmmu_2024}, POPE~\cite{li_evaluating_2023}, AI2D~\cite{kembhavi_diagram_2016}, VizWiz~\cite{gurari_vizwiz_2018}, TextVQA~\cite{singh_towards_2019}, DocVQA~\cite{mathew_docvqa_2021}, ChartQA~\cite{masry_chartqa_2022}, and OCRBench~\cite{Liu_2024}.

\textbf{MME~\cite{fu_mme_2024}.} The MME benchmark is designed to rigorously evaluate a model’s perceptual and cognitive abilities through 14 subtasks. It employs carefully constructed instruction-answer pairs and concise instructions to minimize data leakage and ensure fair evaluation. This setup provides a robust measure of a model’s performance across various tasks. 

\textbf{MMBench~\cite{leonardis_mmbench_2025}.} MMBench offers a hierarchical evaluation framework, categorizing model capabilities into three levels. The first level (L-1) focuses on perception and reasoning. The second level (L-2) expands this to six sub-abilities, while the third level (L-3) further refines these into 20 specific dimensions. This structured approach allows for a nuanced and comprehensive assessment of a model’s multifaceted abilities. 

\textbf{Seed-Bench~\cite{li_seed-bench_2024}.}
SEED-Bench consists of 19K multiple-choice questions with accurate human annotations, covering 12 evaluation dimensions including both the spatial and temporal understanding.

\textbf{GQA~\cite{hudson_gqa_2019}.} GQA is structured around three core components: scene graphs, questions, and images. It includes not only the images themselves but also detailed spatial features and object-level attributes. The questions are crafted to assess a model’s ability to comprehend visual scenes and perform reasoning tasks based on the image content. 

\textbf{ScienceQA~\cite{lu2022learn}.} ScienceQA spans a wide array of domains, including natural, language, and social sciences. Questions are hierarchically categorized into 26 topics, 127 categories, and 379 skills, providing a diverse and comprehensive testbed for evaluating multimodal understanding, multi-step reasoning, and interpretability. 
\begin{table}[t!]
    \centering
    \begin{tabular}{l|ccc}
    \toprule
                     & Stage 1 & Stage 2 & Stage 3 \\
    \midrule
    LLM              &     \includegraphics[width=0.3cm]{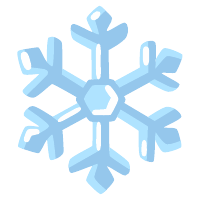}   &    \includegraphics[width=0.3cm]{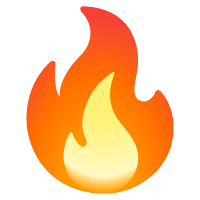}     &  \includegraphics[width=0.3cm]{figures/frozen.pdf}       \\
    Visual Experts   &     \includegraphics[width=0.3cm]{figures/frozen.pdf}    &    \includegraphics[width=0.3cm]{figures/fire.pdf}    &   \includegraphics[width=0.3cm]{figures/frozen.pdf}      \\
    Visual Projector &     \includegraphics[width=0.3cm]{figures/fire.pdf}    &    \includegraphics[width=0.3cm]{figures/fire.pdf}     &     \includegraphics[width=0.3cm]{figures/fire.pdf}    \\
    \midrule
    \tokenselection                 &     -    &    -     &   \includegraphics[width=0.3cm]{figures/fire.pdf}      \\
    \CMOE-LLM                 &      -   &   -      &     \includegraphics[width=0.3cm]{figures/fire.pdf}   \\
    \bottomrule
    \end{tabular}%
    \caption{The training stages of \ours. 
    \tokenselection and \CMOE-LLM are added and fine-tuned in the third stage, while the LLM and vision experts (\CMOE-Vision) remain frozen.
    }
    \label{tab: training stages}
\end{table}

\textbf{MMMU~\cite{yue_mmmu_2024}.}
MMMU includes 11.5K meticulously collected multimodal questions from college exams, quizzes, and textbooks, covering six core disciplines: Art \& Design, Business, Science, Health \& Medicine, Humanities \& Social Science, and Tech \& Engineering. These questions span 30 subjects and 183 subfields, comprising 30 highly heterogeneous image types, such as charts, diagrams, maps, tables, music sheets, and chemical structures.

\textbf{POPE~\cite{li_evaluating_2023}.} POPE is tailored to assess object hallucination in models. It presents a series of binary questions about the presence of objects in images, using accuracy, recall, precision, and F1 score as metrics. This approach offers a precise evaluation of hallucination levels under different sampling strategies. 

\textbf{AI2D~\cite{kembhavi_diagram_2016}.}
AI2D is a dataset of over 5000 grade school science diagrams with over 150000 rich annotations, their ground truth syntactic parses, and more than 15000 corresponding multiple choice questions.

\textbf{VizWiz~\cite{gurari_vizwiz_2018}.}
VizWiz consists of over 31,000 visual questions originating from blind people who each took a picture using a mobile phone and recorded a spoken question about it, together with 10 crowdsourced answers per visual question.

\textbf{TextVQA~\cite{singh_towards_2019}.} TextVQA emphasizes the integration of textual information within images. It evaluates a model’s proficiency in reading and reasoning about text embedded in visual content, requiring both visual and textual comprehension to answer questions accurately. 

\textbf{DocVQA~\cite{mathew_docvqa_2021}.} 
DocVQA consists of 50,000 questions defined on 12,000+ document
images. 

\textbf{ChartQA~\cite{masry_chartqa_2022}.}
CharQA is a large benchmark covering 9.6K human-written
questions as well as 23.1K questions generated from human-written chart summaries.

\textbf{OCRBench~\cite{Liu_2024}.} OCRBench is a comprehensive benchmark for evaluating the OCR capabilities of multi-modal language models across five key tasks: text recognition, scene text-centric and document-oriented VQA, key information extraction, and handwritten mathematical expression recognition.

\subsection{Training Details}

The training of \ours consists of three stages, as shown in \Cref{tab: training stages}. 
\textbf{Stage 1: Warming up for the visual projector}.
This stage pre-trains the visual projector while keeping the LLM and vision experts frozen. 
The LLM and vision experts are initialized from the base models, while the vision projector is randomly initialized. 
\textbf{Stage 2: Supervised fine-tuning}. 
In this stage, we fine-tune the entire model, including the LLM, vision experts, and visual projector. 
\textbf{Stage 3: Supervised fine-tuning for reducing the number of visual tokens}. 
In this stage, we introduce \tokenselection to the model and perform LoRA fine-tuning (\CMOE-LLM) for efficiency. 
Specifically, \tokenselection and \CMOE-LLM are randomly initialized and fine-tuned, while the LLM and vision experts remain frozen. 
To avoid the risk of routing collapse~\cite{shazeer2017}, we employ a balanced loss for \CMOE with a hyperparameter $\lambda = 0.05$, following previous works~\cite{shazeer2017,tian_hydralora_2024}.

\ours uses the same training data as LLaVA-v1.5~\cite{liu2024improvedllava1.5} in stage 3 with 665K instruction data. 
For stage 1 and stage 2, we follow the same data as EAGLE~\cite{shi2024eagle}.
The training was conducted on 8 A6000 GPUs (48 GB) using DeepSpeed’s Zero2 strategy~\cite{10.5555/3433701.3433727}.

We expect the model to learn a comprehensive understanding of the input images in the first two stages, and then in the third stage, the model is guided to focus on the most important information for improved efficiency.

\subsection{Ablation Studies}
\label{sec: ablation: appendix}

\textbf{How do other token reduction methods work?}
To understand how our proposed token reduction module, \textit{i.e.}, \tokenselection, works, we compare it with a representative token reduction method, \textit{i.e.}, MQT-LLaVA~\cite{hu2024matryoshka-MQT-llava}, as shown in \Cref{tab: ablations : MQT}, while keeping the \CMOE unchanged. 
The results show that \tokenselection outperforms MQT on most benchmarks, demonstrating the effectiveness of our proposed method. 
Moreover, compared to MQT-LLaVA (the third row), \CMOE improves the performance of MQT-LLaVA on most benchmarks, showing the importance of introducing multiple experts for extracting informative tokens.

\noindent
\textbf{How does \CMOE-LLM improve the performance?} 
We conduct an ablation study to investigate the effectiveness of \CMOE-LLM in \Cref{tab: ablations : MQT}. 
Results show that \CMOE-LLM significantly improves the performance on most benchmarks, compared to the baseline with \CMOE-Vision and LoRA-MoE~\cite{wu_mixture_2023}. 
\CMOE-LLM improves performance by 62, 0.5, 1.1, 0.6, 0.3, 1.1, 0.5, and 0.7 on MME$^{P}$, SEEDBench$^{I}$, SQA, MMMU, AI2D, TextVQA, ChartQA, and OCRBench, respectively. 
This demonstrates the effectiveness of \CMOE-LLM in better switching between different experts to better understand visual information.

\noindent
\textbf{How does \CMOE-LLM switch between different LORA experts (full results)?}
The full results are presented in \Cref{fig: ablation: expert choice full}. 
The visualization shows that \CMOE-LLM can effectively switch between different experts to better understand visual information. 
We notice that for some layers, such as layers 0, 3, 5, 6, 23, 25, 29, and 31, \CMOE-LLM selects different experts for different benchmarks, while for some layers, such as layers 1, 2, 17, and 27, the model selects the similar experts for different benchmarks.
This demonstrates the effectiveness of \CMOE-LLM in better understanding visual information by switching between different experts.

\noindent
\textbf{How does \CMOE-LLM switch between different LORA experts compared to the vanilla MoE?}
To understand how \CMOE-LLM switches using the additional visual and input signals, we compare it with the vanilla MoE in \Cref{fig: ablation: expert choice comparison}. 
The full routing results of the vanilla MoE are shown in \Cref{fig: ablation: expert choice vanilla MoE}, while the results of \CMOE-LLM are shown in \Cref{fig: ablation: expert choice full}.
The results show that at different layers, \CMOE-LLM shows different experct preferences for different benchmarks, while the vanilla MoE shows similar expert preferences for different benchmarks.

\end{document}